\PassOptionsToPackage{table}{xcolor}
\documentclass[10pt,twocolumn,letterpaper]{article}
\usepackage[pagenumbers]{cvpr}    
\usepackage{algorithm} 
\usepackage[compatible]{algpseudocode} 

\definecolor{cvprblue}{rgb}{0.21,0.49,0.74}
\usepackage[pagebackref,breaklinks,colorlinks,allcolors=cvprblue]{hyperref}

\def\paperID{10606} 
\def\confName{CVPR}
\def\confYear{2025}

\title{ Prototype-Based Image Prompting for Weakly Supervised Histopathological Image Segmentation}

\author{
Qingchen Tang\thanks{Equal contribution.}\quad Lei Fan\footnotemark[1]\quad Maurice Pagnucco\quad Yang Song\\
University of New South Wales\\
{\tt\small \{qingchen.tang@student, lei.fan1@, morri@cse, yang.song1@\}.unsw.edu.au}
}

\begin{document}
\maketitle
\begin{abstract}
Weakly supervised image segmentation with image-level labels has drawn attention due to the high cost of pixel-level annotations. Traditional methods using Class Activation Maps (CAMs) often highlight only the most discriminative regions, leading to incomplete masks. Recent approaches that introduce textual information struggle with histopathological images due to inter-class homogeneity and intra-class heterogeneity. In this paper, we propose a prototype-based image prompting framework for histopathological image segmentation. It constructs an image bank from the training set using clustering, extracting multiple prototype features per class to capture intra-class heterogeneity. By designing a matching loss between input features and class-specific prototypes using contrastive learning, our method addresses inter-class homogeneity and guides the model to generate more accurate CAMs. Experiments on four datasets (LUAD-HistoSeg, BCSS-WSSS, GCSS, and BCSS) show that our method outperforms existing weakly supervised segmentation approaches, setting new benchmarks in histopathological image segmentation.\footnote{\url{https://github.com/QingchenTang/PBIP}}
\end{abstract}

\label{sec:intro}
\setlist[itemize]{left=2pt, label=\textbullet, itemsep=0.5em}

\begin{figure}[t]
  \centering
  \includegraphics[width=\linewidth]{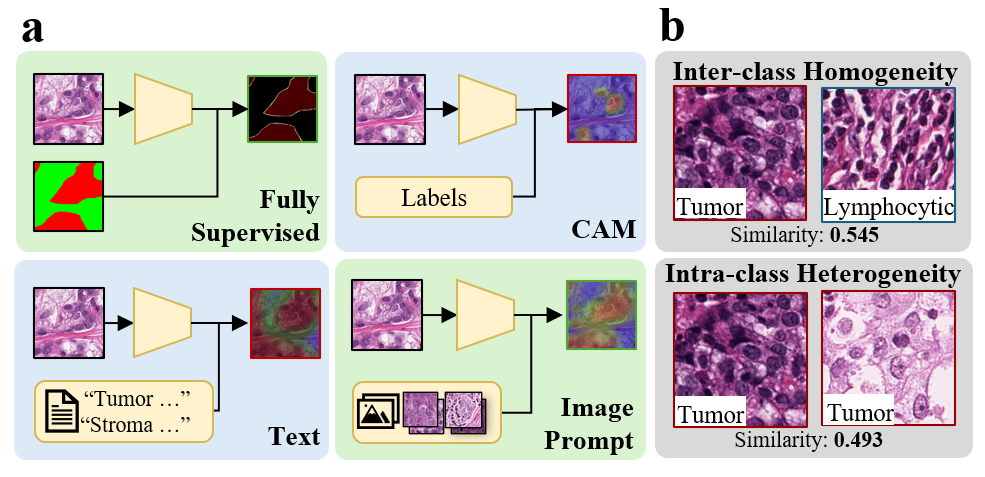}
  \caption{\textbf{a.} Four supervision frameworks for histopathological image segmentation: fully supervised with pixel-level masks, CAM-based WSS using image labels, textural-based WSS, and our image prompt-based framework. \textbf{b.} The challenges of \textit{inter-class homogeneity} (variable texture and staining within classes) and \textit{intra-class heterogeneity} (similar appearances across classes). Cosine similarities are computed using features extracted by the MedCLIP model \cite{wang2022medclip}.}
  \label{fig:overview}
  \vspace{-0.5cm}
\end{figure}

\section{Introduction}
Automated segmentation of histopathology images plays a crucial role in computer-aided diagnosis, assisting in the identification of abnormal tissue regions, quantification of tumor microenvironments~\cite{zhang2024histopathology}, and the support of tumor grading and prognosis~\cite{xu2024whole,fan2021learning,echle2021deep,fan2022cancer,fan2022fast}. Existing methods largely depend on extensive high-quality annotated data~\cite{tajbakhsh2020embracing,liu2021review,minaee2021image}, while pixel-level annotations present particular challenges due to the significant domain expertise and time required for their creation~\cite{fan2025grainbrain,qian2022transformer,zhou2018brief}. To reduce this requirement, weakly supervised segmentation (WSS) frameworks~\cite{kervadec2019constrained,zhang2023tpro,zhang2022transws} have been proposed as a more label-efficient alternative, utilizing weak but easily accessible labels such as bounding boxes~\cite{kervadec2020bounding,qu2024boundary}, scribbles~\cite{liu2022weakly}, point annotations~\cite{qu2020weakly}, and image-level labels~\cite{zhang2023tpro,zhang2022transws}. Typically, WSS methods employ a two-stage process: a classification network is first trained using weak labels to generate pseudo-labels, such as class activation maps (CAMs)~\cite{zhou2016learning}. These pseudo-labels are then utilized as refined supervision signals to train a fully supervised network~\cite{zhou2016learning,han2022multi,zhang2023tpro}. 

However, CAMs generated from image-level labels often focus on the most discriminative regions, leading to incomplete localization of target objects and confusion between target and non-target areas~\cite{lee2022weakly,xie2022clims}. This limitation arises from a domain gap in pre-trained models, which are often trained on source domains (\textit{e.g.}, ImageNet~\cite{deng2009imagenet}) that differ from the target domain (\textit{e.g.}, histopathology), causing CAMs to focus on features misaligned with those of the target domain. Early studies have explored various strategies to improve CAM quality, such as pixel affinity models~\cite{ahn2019weakly} and saliency map integration~\cite{lee2021railroad}. Recent studies have utilized textual information to bridge the gap between image-level labels and pixel-level segmentation~\cite{xie2022clims,deng2024question,zhang2023tpro}. As illustrated in Figure~\ref{fig:overview}, these approaches incorporate textual descriptions (\textit{e.g.}, object type, colour, structure) into the model via pre-trained vision-language models, such as CLIP model~\cite{radford2021learning}. 

While these methods have shown strong performance on popular segmentation benchmarks~\cite{xie2022clims,deng2024question}, they face challenges in histopathological image segmentation. Tissue segmentation involves larger anatomical structures compared to general images or cell nucleus segmentation, with complex spatial arrangements, significant \textit{inter-class homogeneity}, and \textit{intra-class heterogeneity}~\cite{chen2020weakly,li2019weakly}. As illustrated in Figure~\ref{fig:overview}.b, inter-class homogeneity refers to the visual similarity between different tissue types, making it difficult to differentiate target regions from surrounding tissues. For instance, non-tumor tissues may closely resemble tumor tissues, causing CAM-based methods~\cite{chan2019histosegnet,han2022multi} to activate on irrelevant areas erroneously. Intra-class heterogeneity involves significant variability within the same tissue type, such as differences in staining, shape, and texture, which hampers text-based methods~\cite{zhang2023tpro,xie2022clims,deng2024question} from capturing fine-grained details, leading to misalignment between textual descriptions and pixel-level features.

To address these challenges, we propose a Prototype-Based Image Prompting (PBIP) framework for WSS in histopathological segmentation. Unlike methods that rely on text prompts, we utilize image-level labels and their corresponding images to construct visual prototypes. This approach enables the model to discern nuanced visual variations that are challenging for text-based methods. The visual prototypes are achieved through a feature bank that serves as an additional supervisory signal to guide CAM refinement. More specifically, the PBIP framework consists of two main components: a classification network and a prototype-guided feature matching network. The classification network, based on a pre-trained model, extracts multiscale feature representations from the query image to produce pseudo-segmentation masks that indicate class-specific pixel regions. The prototype-guided network constructs a feature bank containing multiple prototypes for each tissue class to capture intra-class heterogeneity. The feature bank is constructed from the training set by organizing images into distinct prototypes based on their labels and employing a CLIP-based image encoder to extract semantically rich, pixel-level features. In addition, a contrastive learning-based similarity loss is computed between the prototypes and class-specific pixel regions, mitigating inter-class homogeneity and yielding high-precision CAMs in the first stage of WSS, which then serve as pseudo-labels for subsequent supervised training.

The main contributions of this paper are as follows: 
\begin{itemize}[nosep] \item To the best of our knowledge, the proposed PBIP framework is the first WSS model in the histopathology domain that utilizes image prompts. By integrating a prototype-guided feature prompting mechanism through an image library, it addresses intra-class heterogeneity and inter-class homogeneity. 
\item We design a CLIP-based image matching method that uses a contrastive learning approach to encourage the alignment of foreground features with their respective prototypes while maintaining distance from non-target prototypes, enhancing pixel-level segmentation in weakly supervised tasks. 
\item Extensive experiments on the LUAD-HistoSeg \cite{han2022multi}, BCSS-wsss \cite{han2022multi}, GCSS \cite{shi2022semi}, and BCSS \cite{amgad2019structured} demonstrate that our method outperforms the current state-of-the-art methods in weakly supervised tissue segmentation. 
\end{itemize}

\section{Related Work}
\label{sec:Related}
\subsection{Weakly Supervised Segmentation}
WSS methods aim to infer pixel-level segmentation from simpler labels such as image-level classification tags~\cite{chen2022class,qian2022transformer}, bounding boxes~\cite{lee2021anti}, or point annotations~\cite{zhang2023weakly,guo2023sac}. Subsequent research has focused on generating higher-quality pseudo labels from these weak annotations, leading to the development of numerous CAM-based approaches~\cite{wang2020score,wang2020self}, such as the Grad-CAM model~\cite{selvaraju2017grad}, pixel affinity models~\cite{ahn2019weakly}, and saliency maps~\cite{lee2021railroad}. These CAM-based models have been used in early histopathological image segmentation studies, including HistoSegNet~\cite{chan2019histosegnet}, OEEN~\cite{li2022online}, and the MLPS model~\cite{han2022multi}.

With advances in multimodal learning \cite{sun2024eggen,wang2024towards}, particularly the development of CLIP~\cite{radford2021learning}, which effectively aligns image and text embeddings, recent studies have incorporated textual information as additional supervision to reduce the discrepancy between image-level labels and pixel-level predictions in WSS. For instance, Xie \textit{et al}. proposed CLIMS~\cite{xie2022clims} and Deng \textit{et al}. introduced QA-CLIMS~\cite{deng2024question}, integrating textual information into WSS models using the pre-trained CLIP model~\cite{radford2021learning}. In histopathological image segmentation, Zhang \textit{et al}. proposed TPRO~\cite{zhang2023tpro}, which incorporates general descriptions of tissue characteristics into image features via an attention mechanism~\cite{vaswani2017attention} to produce more accurate CAMs by aligning image features with textual descriptions. In contrast, our method utilizes image-level labels and the corresponding images to provide finer-grained guidance compared to text-based approaches, addressing the challenges of homogeneity and heterogeneity in histopathological images.

\subsection{Prompt Learning}
Prompt learning initially gained traction in Natural Language Processing (NLP)~\cite{liu2023pre,ding2021openprompt}, where models were guided in specific tasks through designed or learned prompts. With the rise of large-scale pre-trained multimodal models like CLIP~\cite{radford2021learning}, prompt engineering has extended to vision tasks~\cite{zhou2022conditional,zhou2022learning}. Early approaches relied on manually defined text templates (\textit{e.g}., ``a photo of \{class name\}'') to describe classes of interest and extract relevant knowledge~\cite{xie2022clims,deng2024question}. However, this trial-and-error process is time-consuming and requires specific expertise.

To address these limitations, learnable prompt-based methods were introduced for vision tasks~\cite{yang2024foundation,zhou2022conditional}, replacing manual templates with a set of learnable text vectors preceding class names, thus automating the prompt generation process~\cite{zhou2022conditional}. Additionally, some methods attempted to introduce image prompts through prototypes for few-shot image tasks~\cite{hou2022closer,wu2021universal}. For example, Wang \textit{et al}.~\cite{wang2019panet} and Shen \textit{et al}.~\cite{shen2024dual} effectively utilized image prototypes for few-shot image segmentation.

In our method, we leverage prototype-based image prompts instead of text-based prompts. These image prompts offer finer-grained and more direct guidance, allowing the model to capture subtle visual differences that are challenging to describe using text alone.

\begin{figure*}[t]
  \centering
  \includegraphics[width=\textwidth]{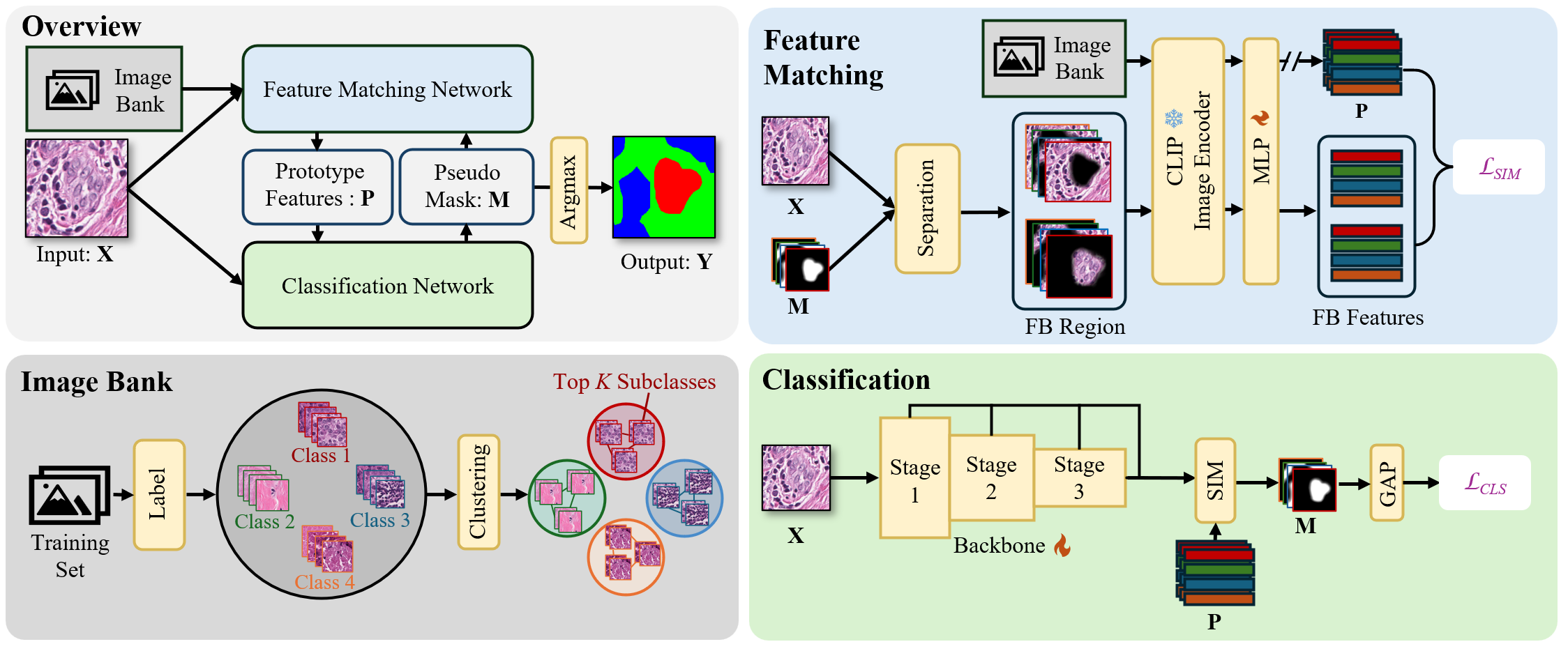}

   \caption{Structure of the proposed PBIP framework. \textbf{Overview.} PBIP consists of two main components: a Classification Network(ClassNet) and an Image Feature Matching Network(ImgMatchNet), which leverage an external image bank to provide image prompts in the form of prototypes. \textbf{Image Bank.} Training images are grouped by their labels and clustered into $K$ subclasses per class. For each subclass, $N_K$ representative images are selected to build the image bank. \textbf{ClassNet.} It receives an input image \(\mathbf{X}\) and prototype features \(\mathbf{P}\), performing a classification task to generate the pseudo-segmentation mask \(\mathbf{M}\). \textbf{ImgMatchNet.} It processes the input image \(\mathbf{X}\) and the initial pseudo-segmentation mask \(\mathbf{M}\), extracting foreground and background regions. These regions are then matched with \(\mathbf{P}\) from the image bank to refine the pseudo mask generation.}
   \label{fig:IMSH2}
   \vspace{-0.4cm}
\end{figure*}

\section{Method}
\label{sec:Method}
\subsection{Overview}
The key idea of our proposed PBIP framework is to leverage image-level labels and corresponding images to create a high-quality visual prototype feature bank, guiding the generation of more accurate CAMs as pseudo masks for WSS. As illustrated in Figure~\ref{fig:IMSH2}, the PBIP framework consists of two main components: a Classification Network (ClassNet) and an Image Feature Matching Network (ImgMatchNet).

The ClassNet takes a histopathology image $\mathbf{X} \in \mathbb{R}^{H \times W \times C}$ from the dataset and the prototype feature vectors \( \mathbf{P}_i\) from ImgMatchNet (as detailed in Section~\ref{sec:Method.3}), where $H$, $W$, and $C$ denote the height, width, and number of channels, respectively. ClassNet performs a classification task to generate initial pseudo-segmentation masks $\mathbf{M} \in \mathbb{R}^{H \times W \times N}$ and is optimized by a classification loss $\mathcal{L}_{\text{CLS}}$, where $N$ denotes the number of classes in the dataset.

The ImgMatchNet receives the input histopathology image $\mathbf{X}$, an image bank constructed from the training set through clustering, and the pseudo-segmentation masks \(\mathbf{M}\) generated by the ClassNet. ImgMatchNet extracts prototype features \(\mathbf{P}\) using a CLIP image encoder from the image bank and refines the pseudo-segmentation masks \(\mathbf{M}\) through a similarity matching loss $\mathcal{L}_{\text{SIM}}$. The refined masks \(\mathbf{M}\) are merged via an argmax operation over the channel dimension to produce the final activation map for the first stage of WSS. These masks are subsequently used to train a fully supervised segmentation model in the second stage, generating the final segmentation masks.

\subsection{Construction of Image Bank}
\label{sec:Method.1}
We construct an image bank that captures diverse visual prototypes for each class, which is automatically assembled from the training set. Specifically, we select tissue images based on image-level labels that indicate the patch contains only one class (e.g., only tumor) and exclude images with excessive white regions automatically. These patches are grouped into $N$ categories corresponding to the $N$ classes in the dataset. To capture intra-class heterogeneity, $K$-Means clustering is applied within each category to partition the patches into $K$ subcategories. For each subcategory, the top $N_K$ images closest to the cluster center are selected as prototypes for the image bank. The distance metric used for clustering is defined as:
\begin{equation} 
D_t(x_1, x_2) = 1 - \frac{\phi_e(x_1) \cdot \phi_e(x_2)}{\|\phi_e(x_1)\| \|\phi_e(x_2)\|}, 
\end{equation}
where $x_1$ and $x_2$ are two image patches, $\phi_e$ denotes the feature extraction function of the CLIP image encoder, and the symbol $\cdot$ denotes the dot product operation between vectors. This process results in an image bank comprising $N$ classes, each containing $K$ subcategories, with each subcategory holding $N_K$ representative images.

\subsection{Classification Network}
\label{sec:Method.2}

The ClassNet generates the initial pseudo-segmentation masks \(\mathbf{M}\) by leveraging both the input image \( \mathbf{X} \) and the prototype features \( \mathbf{P} \). We adopt SegFormer~\cite{xie2021segformer} as the backbone due to its efficacy in capturing multi-scale contextual information through its hierarchical transformer encoder.

Given an input histopathology image \( \mathbf{X} \), hierarchical feature maps \( \mathbf{F}_i \in \mathbb{R}^{(H/2^{i+1}) \times (W/2^{i+1}) \times C_i} \) are extracted across different stages \( i = 1, 2, 3 \), with \( C_{i+1} > C_i \), where $C_i$ corresponds to the number of channels at each hierarchical level $i$ of the feature maps. To generate the initial pseudo masks, the cosine similarity is computed (\textit{SIM}) between each pixel feature vector in \( \mathbf{F}_i \) and the prototype feature vectors in \( \mathbf{P}_i \in \mathbb{R}^{N \times K \times C_i}\). This computation produces confidence scores for each pixel across all classes, forming the pseudo-segmentation masks \( \mathbf{M}_i \in \mathbb{R}^{(H/2^{i+1}) \times (W/2^{i+1}) \times N} \). The confidence score for pixel \( p \) and class \( n \) at level \( i \) is calculated as:
\begin{equation} 
\mathbf{M}_i(p, n) = \frac{1}{K} \sum_{k=1}^{K} \frac{ \mathbf{F}_i(p) \cdot \mathbf{P}_i(n, k) }{ \|\mathbf{F}_i(p) \|\ \mathbf{P}_i(n, k) \| }, 
\end{equation}
where \( \mathbf{F}_i(p) \) is the feature vector of pixel \( p \) in \( \mathbf{F}_i \), and \( \mathbf{P}_i(n, k) \) is the \( k \)-th prototype feature vector for class \( n \). By averaging the cosine similarities over all prototypes \( K \) for each class, it effectively captures intra-class variations and obtains a robust confidence score that reflects the likelihood of the pixel \( p \) belonging to the class \( n \).

\subsection{Image Feature Matching Network}
\label{sec:Method.3}

ImgMatchNet introduces image prompts through a prototype-based approach. It generates prototype features \(\mathbf{P}\) from the image bank, and then ClassNet is used to compute pixel-level class confidences in the feature maps. To extract these prototype features, we employ the image encoder from the pre-trained MedCLIP model~\cite{wang2022medclip}, a variant of the CLIP architecture tailored for medical images and trained on large-scale medical image-text pairs.

All images in the image bank are encoded using the MedCLIP image encoder, producing feature representations denoted as \( \mathbf{F}_p \in \mathbb{R}^{N \times K \times N_K \times d} \), where \( N \) is the number of classes in the dataset, \( K \) is the number of subclasses per class determined by the clustering, \( N_K \) is the number of images per subclass, and \( d \) is the dimension of the feature space. The mean feature vector is computed for each subclass across its images to obtain the prototype features, yielding \( \mathbf{P} \in \mathbb{R}^{N \times K \times d} \). To align these prototype features with the hierarchical feature maps generated by the ClassNet, we employ a Multi-Layer Perceptron (MLP) composed of fully connected layers and ReLU activations. The MLP projects the prototype features to match the dimensionality of each feature level, producing \( \mathbf{P}_i \in \mathbb{R}^{N \times K \times C_i} \).

ImgMatchNet includes a foreground-background separation module to further refine the pseudo-masks obtained from the ClassNet. Specifically, this module upsamples and aggregates the hierarchical pseudo-masks from different feature levels to obtain a comprehensive pseudo-mask:
\begin{equation}
\mathbf{M'} = \sum_{i=1}^{3} \text{Up}(\mathbf{M} _i),
\end{equation}
where $\mathbf{M'} \in \mathbb{R}^{H \times W \times N}$, and $\text{Up}(\cdot)$ denotes the upsampling operation to match the original image resolution.

We employ an adaptive thresholding module to separate foreground and background regions. This module computes an adaptive threshold $\tau$ based on the intensity distribution of the pseudo-mask, dynamically distinguishing between foreground and background to reduce noise. The adaptive threshold is computed as:
\begin{equation}
\tau = \delta \cdot \max (\mathbf{M}), \quad \delta \in [0, 1],
\end{equation}
where $\delta$ is a scaling parameter that controls the threshold level. Applying this threshold to the activation map yields a binary mask $b \in \mathbb{R}^{H \times W \times N}$, where pixel values greater than or equal to $\tau$ are set to 1 (foreground), and values below $\tau$ are set to 0 (background). We then separate foreground and background images through element-wise multiplication:
\begin{equation}
\mathbf{X}_{\text{FG}} = b \cdot \mathbf{M} \cdot \mathbf{X}, \quad \mathbf{X}_{\text{BG}} = (1 - b) \cdot (1 - \mathbf{M}) \cdot\mathbf{X},
\end{equation}
where $\mathbf{X}_{\text{FG}}, \mathbf{X}_{\text{BG}} \in \mathbb{R}^{H \times W \times N}$ represent the separated foreground and background regions. By using the same MedCLIP image encoder and MLP, we extract hierarchical foreground and background features $\mathbf{F}_{\text{FG}_i}, \mathbf{F}_{\text{BG}_i} \in \mathbb{R}^{N \times C_i}$. These features are then utilized in a similarity matching loss to optimize the generation of the pseudo-masks further.

\subsection{Optimization Objectives}
Our model is jointly optimized by two objectives: the classification loss $\mathcal{L}_{\text{CLS}}$, which ensures consistency between predicted labels and image-level annotations, and the similarity loss $\mathcal{L}_{\text{SIM}}$, which promotes alignment between the encoded features of the foreground and background regions with their respective prototype features \(\mathbf{P}\). The total loss function is defined as:
\begin{equation}
\mathcal{L}_{\text{total}} = \alpha \cdot \mathcal{L}_{\text{CLS}} + \beta \cdot \mathcal{L}_{\text{SIM}},
\end{equation}
where $\alpha$ and $\beta$ are weighting coefficients that balance the contribution of each loss term.

\textbf{Classification Loss.} Image-level class predictions $\hat{y} \in \mathbb{R}^{3 \times N}$ are obtained by applying global average pooling to the pseudo-segmentation masks \(\mathbf{M}\) from each hierarchical level in the ClassNet, where $3$ corresponds to the levels and $N$ is the number of classes. The predictions are compared with the ground-truth labels $y \in \mathbb{R}^{1 \times N}$, and the classification loss for each level $i$ is:
\begin{equation}
\mathcal{L}_{\text{CLS}}[i] = \textit{CossEntropy}(y, \hat{y}[i])
\end{equation}
where $\sigma$ is the sigmoid function. The total classification loss is a weighted sum:
\begin{equation}
\mathcal{L}_{\text{CLS}} = \sum_{i=1}^{3} \mathcal{L}_{\text{CLS}}[i].
\end{equation}

\textbf{Similarity Loss.} The similarity loss \(\mathcal{L}_{\text{SIM}}\) is divided into two components: the foreground similarity loss \(\mathcal{L}_{\text{FGS}}\) and the background similarity loss \(\mathcal{L}_{\text{BGS}}\).

The foreground similarity loss $\mathcal{L}_{\text{FGS}}$ aligns the foreground image features with the corresponding foreground prototype features, while distinguishing them from the background prototype features. Formally, given the foreground image feature $F_{\text{FG}}[j]$ for the $j$-th class, foreground prototype feature $P_\text{FG}=P[j]$ , and the background prototype features $P_\text{BG}=P[m]$ where $1\leq m\leq N$ and $m \neq j$, the loss is defined as:

\begin{equation}
\mathcal{L}_{\text{FGS}} = -\log \left( \frac{\exp \left( {\mathbf{s}_{\text{j}}^{\text{FF}}}/{\tau} \right)}{\exp \left( {\mathbf{s}_{\text{j}}^{\text{FF}}}/{\tau} \right) + \exp \left( {\mathbf{s}_{\text{j}}^{\text{FB}}}/{\tau} \right)} \right),
\end{equation}
where
\begin{equation}
\mathbf{s}_{\text{j}}^{\text{FF}}=sum(F_{\text{FG}}[j] \cdot P_\text{FG}), \mathbf{s}_{\text{j}}^{\text{FB}}=sum(F_{\text{FG}}[j] \cdot P_\text{BG}),
\end{equation}
where $\mathbf{s}_{\text{j}}^{\text{FF}}$ is the sum of similarity scores between the foreground feature and the foreground prototype of the $j$-th class, while $\mathbf{s}_{\text{j}}^{\text{FB}}$ is the sum of similarity scores between the foreground feature of the $j$-th class and the background prototype. The temperature parameter \( \tau \) controls the concentration level of the distribution.

Similarly, \(\mathcal{L}_{\text{BGS}}\) aligns background image features with background prototype features while distinguishing them from foreground prototype features. Given the background image feature $F_{\text{BG}}[j]$, the background prototype feature $P_{\text{BG}}=P[j]$ for the $j$-th region, and the foreground prototype features $P_{\text{FG}}=P[m]$, the loss is defined as:

\begin{equation}
\mathcal{L}_{\text{BGS}} = -\log \left( \frac{\exp \left( {\bar{\mathbf{s}}_{\text{j}}^{\text{BB}}}/{\tau} \right)}{\exp \left( {\bar{\mathbf{s}}_{\text{j,j}}^{\text{BF}}}/{\tau} \right) + \exp \left( {\bar{\mathbf{s}}_{\text{j}}^{\text{BB}}}/{\tau} \right)} \right),
\end{equation}

\begin{equation}
\bar{\mathbf{s}}_{\text{j}}^{\text{BB}}=mean(F_{\text{BG}}[j] \cdot P_\text{BG}),
\end{equation}

where $\bar{\mathbf{s}}_{\text{j}}^{\text{BB}}$ is the mean of similarity scores between the background feature of the $j$-th class and the background prototype. The total similarity loss is the combination of the two components:
\begin{equation}
\mathcal{L}_{\text{SIM}} = \theta_1\cdot \mathcal{L}_{\text{FGS}} + \theta_2 \cdot \mathcal{L}_{\text{BGS}}.
\end{equation}
where \(\theta_1\) and \(\theta_2\) are weights that balance the importance of the foreground and background similarity terms. 

\section{Experiments}
\begin{table*}[htbp]
\centering
\resizebox{\textwidth}{!}{
\begin{tabular}{l|l|llll|llll}
\hline
  & Dataset              & \multicolumn{4}{c|}{BCSS-WSSS~\cite{han2022multi}}           & \multicolumn{4}{c}{LUAD-HistoSeg~\cite{han2022multi}} \\ \hline
Sup.          & Method               & mIoU(\%)     & FwIoU(\%)     & bIoU(\%)      & dice(\%)      & mIoU(\%)      & FwIoU(\%)     & bIoU(\%)         & dice(\%) \\ \hline 
              & HistoSegNet~\cite{chan2019histosegnet} & 32.19\scalebox{0.7}{$\pm$5.8} & 35.69\scalebox{0.7}{$\pm$5.9} & 19.24\scalebox{0.7}{$\pm$1.7} & 48.70\scalebox{0.7}{$\pm$6.8} & 45.32\scalebox{0.7}{$\pm$4.3} & 45.08\scalebox{0.7}{$\pm$3.9} & 18.36\scalebox{0.7}{$\pm$1.1} & 62.37\scalebox{0.7}{$\pm$4.3} \\
              & TransWS~\cite{zhang2022transws} & 40.25\scalebox{0.7}{$\pm$0.0} & \multicolumn{1}{c}{-} & \multicolumn{1}{c}{-} & 57.39\scalebox{0.7}{$\pm$0.0} & 56.27\scalebox{0.7}{$\pm$0.0} & \multicolumn{1}{c}{-} & \multicolumn{1}{c}{-} & 72.02\scalebox{0.7}{$\pm$0.0} \\
              & OEEN~\cite{li2022online} & 59.93\scalebox{0.7}{$\pm$0.5} & 61.11\scalebox{0.7}{$\pm$0.8} & 25.09\scalebox{0.7}{$\pm$1.2} & 74.95\scalebox{0.7}{$\pm$0.4} & 72.54\scalebox{0.7}{$\pm$0.2} & 71.65\scalebox{0.7}{$\pm$0.2} & 26.15\scalebox{0.7}{$\pm$1.4} & 84.08\scalebox{0.7}{$\pm$0.2} \\
$\mathcal{W}$ & MLPS~\cite{han2022multi} & 65.17\scalebox{0.7}{$\pm$1.7} & 69.65\scalebox{0.7}{$\pm$1.9} & 35.19\scalebox{0.7}{$\pm$0.5} & 78.91\scalebox{0.7}{$\pm$1.1} & 73.54\scalebox{0.7}{$\pm$2.1} & 73.15\scalebox{0.7}{$\pm$1.1} & 27.77\scalebox{0.7}{$\pm$0.9} & 84.75\scalebox{0.7}{$\pm$1.4} \\
              & TPRO~\cite{zhang2023tpro} & 63.92\scalebox{0.7}{$\pm$0.4} & 64.45\scalebox{0.7}{$\pm$0.2} & 25.79\scalebox{0.7}{$\pm$0.5} & 77.99\scalebox{0.7}{$\pm$0.3} & 74.14\scalebox{0.7}{$\pm$0.8} & 74.01\scalebox{0.7}{$\pm$0.2} & 27.46\scalebox{0.7}{$\pm$0.3} & 85.14\scalebox{0.7}{$\pm$0.5} \\
              & CLIMS~\cite{xie2022clims} & 38.98\scalebox{0.7}{$\pm$2.4} & 37.58\scalebox{0.7}{$\pm$2.1} & 13.52\scalebox{0.7}{$\pm$0.9} & 56.09\scalebox{0.7}{$\pm$2.6} & 39.45\scalebox{0.7}{$\pm$3.1} & 38.92\scalebox{0.7}{$\pm$2.7} & 17.55\scalebox{0.7}{$\pm$1.6} & 56.58\scalebox{0.7}{$\pm$3.0} \\
              & QA-CLIMS~\cite{deng2024question} & 22.45\scalebox{0.7}{$\pm$9.7} & 24.87\scalebox{0.7}{$\pm$8.2} & 11.15\scalebox{0.7}{$\pm$2.4} & 36.67\scalebox{0.7}{$\pm$11.2} & 21.97\scalebox{0.7}{$\pm$8.6} & 20.63\scalebox{0.7}{$\pm$8.9} & 10.67\scalebox{0.7}{$\pm$2.1} & 36.03\scalebox{0.7}{$\pm$10.0} \\
              & Proto2Seg~\cite{pan2023human}    & 67.86\scalebox{0.7}{$\pm$0.9$^*$}    & 72.48\scalebox{0.7}{$\pm$1.1$^*$}   & 35.54\scalebox{0.7}{$\pm$0.4$^*$}  & 81.00\scalebox{0.7}{$\pm$0.9$^*$}   & 74.91\scalebox{0.7}{$\pm$1.2$^*$}     & 74.52\scalebox{0.7}{$\pm$1.0$^*$}   & 30.71\scalebox{0.7}{$\pm$0.5$^*$}   & 85.97\scalebox{0.7}{$\pm$1.2$^*$}  \\\cline{2-10} 
              & PBLP (Ours)  & \textbf{69.54\scalebox{0.7}{$\pm$0.6}} & \textbf{74.52\scalebox{0.7}{$\pm$0.5}} & \textbf{37.17\scalebox{0.7}{$\pm$0.4}} & \textbf{82.03\scalebox{0.7}{$\pm$0.5}} & \textbf{76.44\scalebox{0.7}{$\pm$0.5}} & \textbf{75.19\scalebox{0.7}{$\pm$0.5}} & \textbf{30.64\scalebox{0.7}{$\pm$0.3}} & \textbf{86.64\scalebox{0.7}{$\pm$0.3}} \\ \hline
\hline
& Dataset              & \multicolumn{4}{c|}{GCSS~\cite{shi2022semi}}           & \multicolumn{4}{c}{BCSS~\cite{amgad2019structured}} \\ \hline 
Sup.          & Method               & mIoU(\%)     & FwIoU(\%)     & bIoU(\%)      & dice(\%)      & mIoU(\%)      & FwIoU(\%)     & bIoU(\%)         & dice(\%) \\ \hline
\rowcolor[HTML]{EFEFEF} 
 & SSPCL~\cite{shi2022semi} & 51.63\scalebox{0.7}{$\pm$0.0}    & \multicolumn{1}{c}{-}           & \multicolumn{1}{c}{-}           & 68.10\scalebox{0.7}{$\pm$0.0}   & 62.47\scalebox{0.7}{$\pm$0.0}    & \multicolumn{1}{c}{-}           & \multicolumn{1}{c}{-}          & 76.90\scalebox{0.7}{$\pm$0.0} \\
 \rowcolor[HTML]{EFEFEF} 
$\mathcal{F \& S}$ & SAM-Path~\cite{zhang2023sam} & 58.78\scalebox{0.7}{$\pm$0.9} & 59.01\scalebox{0.7}{$\pm$0.7} & 58.78\scalebox{0.7}{$\pm$0.9} & 73.11\scalebox{0.7}{$\pm$0.3} & 63.27\scalebox{0.7}{$\pm$0.8}    & 63.35\scalebox{0.7}{$\pm$0.7}   & 25.10\scalebox{0.7}{$\pm$0.7}   & 77.50\scalebox{0.7}{$\pm$0.6} \\ 
\rowcolor[HTML]{EFEFEF} 
              & SAM2-Path~\cite{zhang2024sam2}  & 57.06\scalebox{0.7}{$\pm$0.6}    & 56.50\scalebox{0.7}{$\pm$0.6}   & 22.32\scalebox{0.7}{$\pm$0.4}   & 72.66\scalebox{0.7}{$\pm$0.7}   & 61.57\scalebox{0.7}{$\pm$0.6}   & 62.01\scalebox{0.7}{$\pm$0.6}    & 23.77\scalebox{0.7}{$\pm$0.4}   & 76.21\scalebox{0.7}{$\pm$0.5} \\ \hline
              & HistoSegNet~\cite{chan2019histosegnet} & 30.02\scalebox{0.7}{$\pm$6.2}    & 31.25\scalebox{0.7}{$\pm$5.4}   & 10.10\scalebox{0.7}{$\pm$3.6}   & 46.15\scalebox{0.7}{$\pm$7.0}   & 22.15\scalebox{0.7}{$\pm$7.8}    & 22.80\scalebox{0.7}{$\pm$8.0}   & 12.11\scalebox{0.7}{$\pm$4.1}   & 36.27\scalebox{0.7}{$\pm$8.9} \\
              & OEEN~\cite{li2022online}     & 45.89\scalebox{0.7}{$\pm$2.8}    & 47.14\scalebox{0.7}{$\pm$2.5}   & 15.33\scalebox{0.7}{$\pm$1.8}   & 62.91\scalebox{0.7}{$\pm$3.1}   & 47.50\scalebox{0.7}{$\pm$0.4}    & 48.21\scalebox{0.7}{$\pm$0.3}   & 14.00\scalebox{0.7}{$\pm$0.2}   & 64.41\scalebox{0.7}{$\pm$0.4} \\
$\mathcal{W}$ & MLPS~\cite{han2022multi}     & 51.20\scalebox{0.7}{$\pm$1.7}    & 52.85\scalebox{0.7}{$\pm$1.5}   & 19.10\scalebox{0.7}{$\pm$1.3}   & 67.72\scalebox{0.7}{$\pm$2.4}   & 49.32\scalebox{0.7}{$\pm$1.1}    & 51.31\scalebox{0.7}{$\pm$1.3}    & 21.08\scalebox{0.7}{$\pm$0.7}    & 66.06\scalebox{0.7}{$\pm$0.9} \\
              & TPRO~\cite{zhang2023tpro}    & 50.12\scalebox{0.7}{$\pm$1.9}    & 51.73\scalebox{0.7}{$\pm$1.7}   & 18.50\scalebox{0.7}{$\pm$1.1}   & 69.84\scalebox{0.7}{$\pm$1.8}   & 50.97\scalebox{0.7}{$\pm$0.3}     & 51.01\scalebox{0.7}{$\pm$0.3}   & 17.88\scalebox{0.7}{$\pm$0.4}   & 67.52\scalebox{0.7}{$\pm$0.2}  \\
              & CLIMS~\cite{xie2022clims}    & 32.03\scalebox{0.7}{$\pm$3.0}    & 31.89\scalebox{0.7}{$\pm$2.7}   & 15.22\scalebox{0.7}{$\pm$1.8}   & 48.52\scalebox{0.7}{$\pm$3.3}   & 36.12\scalebox{0.7}{$\pm$4.6}    & 36.77\scalebox{0.7}{$\pm$3.0}   & 12.16\scalebox{0.7}{$\pm$1.1}   & 53.07\scalebox{0.7}{$\pm$4.2} \\
              & QA-CLIMS~\cite{deng2024question} & 25.01\scalebox{0.7}{$\pm$8.3}    & 26.54\scalebox{0.7}{$\pm$7.6}   & 12.34\scalebox{0.7}{$\pm$4.2}   & 40.02\scalebox{0.7}{$\pm$9.0}   & 23.00\scalebox{0.7}{$\pm$9.9}    & 24.07\scalebox{0.7}{$\pm$9.1}   & 10.27\scalebox{0.7}{$\pm$3.6}   & 37.39\scalebox{0.7}{$\pm$12.2} \\
              & Proto2Seg~\cite{pan2023human}    & 53.23\scalebox{0.7}{$\pm$2.8}    & 53.97\scalebox{0.7}{$\pm$2.9}   & \textbf{22.34\scalebox{0.7}{$\pm$1.7}}   & 70.84\scalebox{0.7}{$\pm$2.0}   & \textbf{55.34\scalebox{0.7}{$\pm$2.0}}     & \textbf{56.72\scalebox{0.7}{$\pm$2.0}}   & 22.40\scalebox{0.7}{$\pm$1.5}   & 70.78\scalebox{0.7}{$\pm$2.1} \\\cline{2-10} 
              & PBLP (Ours)        & \textbf{55.10\scalebox{0.7}{$\pm$1.3}}   & \textbf{57.05\scalebox{0.7}{$\pm$1.2}}   & 21.40\scalebox{0.7}{$\pm$0.9}   & \textbf{71.05\scalebox{0.7}{$\pm$1.4}}   & 55.28\scalebox{0.7}{$\pm$0.8}   & 56.30\scalebox{0.7}{$\pm$0.8}   & \textbf{23.65\scalebox{0.7}{$\pm$0.3}}   & \textbf{71.20\scalebox{0.7}{$\pm$0.7}} \\ \hline
\end{tabular}
}
\caption{Quantitative comparison of various methods on four histopathological datasets. The best results are highlighted in bold. “Sup.” denotes the type of supervision used: $\mathcal{F \& S}$ represents fully and semi-supervised methods, while $\mathcal{W}$ indicates weakly supervised methods. Except for CLIMS and QA-CLIMS (designed for natural image segmentation), all methods are specifically tailored for histopathological image segmentation. For the BCSS-WSSS and LUAD-HistoSeg, the $p$-values from t-tests comparing our model with the second-best results$^*$ were calculated, and all $p$-values were less than 0.05.}
\label{tab:1}
\vspace{-0.3cm}
\end{table*}

\begin{table}[htbp]
\centering
\resizebox{0.48\textwidth}{!}{
\begin{tabular}{ccc|cccc}
\hline
$\mathcal{L}_{CLS}$     & $\mathcal{L}_{FGM}$ & $\mathcal{L}_{BGM}$  & mIoU(\%) & FwIoU(\%) & bIoU(\%)  & dice(\%)      \\ \hline
\checkmark              &                     &                      & 62.08±0.3& 63.00±0.4 & 25.73±0.2 & 76.60±0.3           \\
\checkmark              & \checkmark          &                      & 55.01±2.8& 59.47±2.5 & 20.44±1.0 & 70.98±3.0        \\
\checkmark              &                     &\checkmark            & 50.23±2.4& 52.11±1.7 & 19.20±0.7 & 66.87±2.7          \\
                        & \checkmark          &\checkmark            & 62.34±1.6& 64.08±1.9 & 25.68±0.7 & 76.80±1.5           \\
\checkmark              & \checkmark          & \checkmark           & \textbf{67.54±0.6}& \textbf{69.42±0.7} & \textbf{33.00±0.4} & \textbf{80.63±0.4}          \\ \hline
\end{tabular}
}
\caption{Ablation study of different loss function combinations on the initial pseudo masks quality (mIoU\%) for the BCSS-WSSS dataset. }
\label{tab:4}
\vspace{-0.5cm}
\end{table}

\subsection{Experimental Setup}
\textbf{Datasets.} We evaluated our method on four histopathological datasets. \textbf{BCSS}~\cite{amgad2019structured} consists of 151 H\&E-stained breast cancer images from TCGA-BRCA, annotated with four classes (Tumor, Stroma, Lymphocytic infiltrate, Necrosis), and includes 30,000 training patches, 2,500 validation patches and 2,500 testing patches. 

\textbf{BCSS-WSSS}~\cite{han2022multi} is derived from BCSS for WSS with 23,422 training patches, 3,418 validation patches and 4,986 testing patches. 

\textbf{LUAD-HistoSeg}~\cite{han2022multi} contains 17,291 H\&E-stained lung adenocarcinoma patches, annotated with four classes (Tumor Epithelium, Stroma, Necrosis, and Lymphocytes), with 16,678 training patches, 306 validation patches and 307 testing patches. 

\textbf{GCSS}~\cite{shi2022semi} comprises 100 H\&E-stained Gastric cancer images from TCGA-STAD gastric cancer images, annotated with four classes (Tumor, Lymphoid Stroma, Desmoplastic Stroma, Smooth Muscle Necrosis) providing 20,000 training, 2,500 validation patches and 2,500 testing patches.

BCSS-WSSS and LUAD-HistoSeg are used exclusively for weakly supervised segmentation tasks, as the training images are labelled only at the image level, with the validation and test sets still including ground truth masks for evaluation purposes. More details about datasets are in the Supplementary Material.

\textbf{Implementation Details.} The ClassNet utilizes the Mix Transformer from SegFormer~\cite{xie2021segformer} as its backbone, which is pretrained on ImageNet-1K. The ImgMatchNet is built upon an image bank assembled from the training set, where the number of clusters $K$ in $K$-Means was set to 3, and each subclass contains \( N_K \)=100. The scaling parameter $\delta$ in the adaptive thresholding module was 0.15.

During training, we utilized the AdamW optimizer with an initial learning rate of $1 \times 10^{-5}$ and a weight decay of 0.003. The total loss function combines the classification loss and similarity loss with weighting factors $\alpha = 1$ and $\beta = 0.5$, respectively. The temperature parameter in both \(\mathcal{L}_{\text{FGS}}\) and \(\mathcal{L}_{\text{BGS}}\) was set to 1, while $\theta_1$ and $\theta_2$ were set to 1 and 0.5, respectively. The model was trained for 10 epochs, with a batch size of 10 during the first stage. In the second stage in WSS, an unmodified Deeplab-v2 model~\cite{chen2017deeplab} was trained as a fully supervised segmentation model. All models were trained on a single 4090 GPU.

To quantitatively evaluate our method, we employed four standard metrics: Mean Intersection over Union (mIoU), Frequency Weighted IoU (FWIoU), Boundary IoU (bIoU), and Dice Coefficient. We also calculated p-values from t-tests to indicate the statistical significance of our results, measuring the probability that the observed differences occurred by chance.

\subsection{Results}

\begin{figure}[ht]
  \centering
  \begin{subfigure}{0.48\linewidth}
    \includegraphics[width=\textwidth]{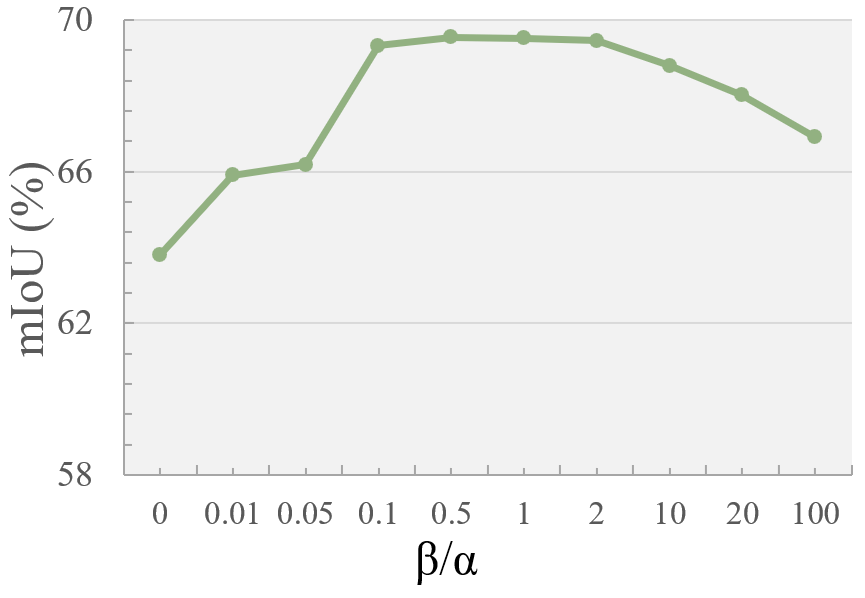}
    \caption{Analysis of $\beta/\alpha$.}
    \label{fig:short-a}
  \end{subfigure}
  \hfill
  \begin{subfigure}{0.49\linewidth}
    \includegraphics[width=\textwidth]{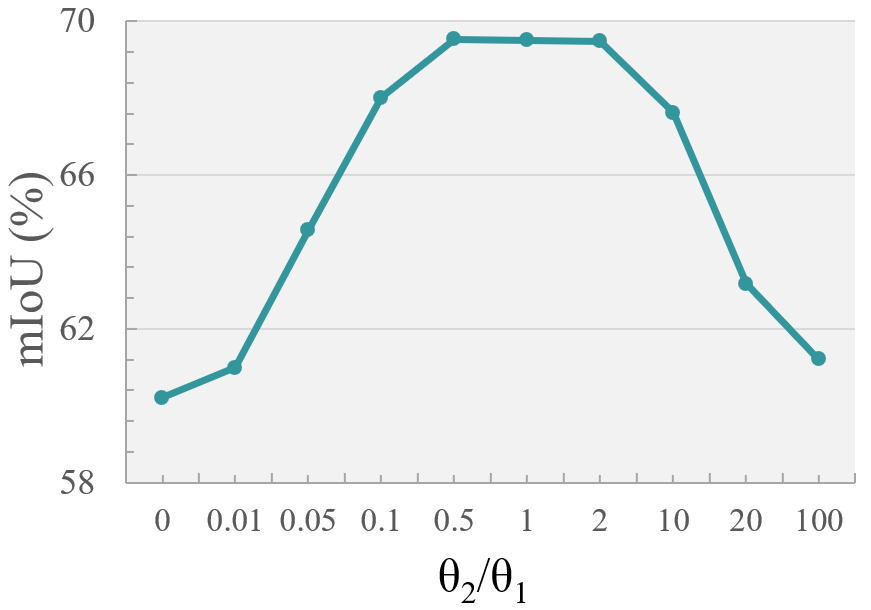}
    \caption{Analysis of $\theta_2/\theta_1$.}
    \label{fig:short-b}
  \end{subfigure}
  \caption{Ablation study on hyperparameter ratios. The mIoU values are reported on initial pseudo masks for BCSS-WSSS.}

  \label{fig:hp}
\end{figure}

\begin{table}[t]
\resizebox{0.48\textwidth}{!}{
\begin{tabular}{ll|llll}
\hline
\multicolumn{2}{l|}{Dataset}            & \multicolumn{4}{l}{BCSS}               \\ \hline
\multicolumn{1}{l|}{Encoder}            & Backbone  & mIoU(\%)      & FwIoU(\%)    & bIoU(\%)      & dice(\%)  \\ \hline
\multicolumn{1}{l|}{}                   & SegFormer & 66.45±0.3&72.42±0.3& 27.05±0.2&80.56±0.2    \\
\multicolumn{1}{l|}{}                   & ResNet18  & 65.87±0.2&70.40±0.3&23.23±0.3 &78.95±0.2    \\
\multicolumn{1}{l|}{{CLIP~\cite{radford2021learning}}}    & ResNet38  & 66.10±0.4&71.22±0.4&26.56±0.2 &79.86±0.3    \\
\multicolumn{1}{l|}{}                   & ResNet50  & 65.30±0.4&70.21±0.4&23.31±0.3 &78.27±0.3    \\
\multicolumn{1}{l|}{}                   & TransUNet & 60.17±0.4&63.52±0.5&17.81±0.2 &75.12±0.5    \\ \hline
\multicolumn{1}{l|}{}                   & SegFormer & \textbf{68.01±0.6}&73.20±0.5&36.88±0.4&81.62±0.5     \\
\multicolumn{1}{l|}{}                   & ResNet18  & 67.17±0.4&72.60±0.3&33.99±0.3&80.64±0.4    \\
\multicolumn{1}{l|}{{MedCLIP~\cite{wang2022medclip}}} & ResNet38  & 67.18±0.5&72.57±0.5&35.18±0.3&80.93±0.4    \\
\multicolumn{1}{l|}{}                   & ResNet50  & 67.03±0.3&72.11±0.2&33.98±0.4&80.26±0.2    \\
\multicolumn{1}{l|}{}                   & TransUNet & 62.09±0.5&66.89±0.5&20.15±0.3&76.61±0.5  \\ \hline
\multicolumn{1}{l|}{}                   & SegFormer & 67.16±0.7&\textbf{73.53±0.5}&\textbf{37.21±0.4}&\textbf{81.77±0.5}    \\
\multicolumn{1}{l|}{}                   & ResNet18  & 66.89±0.6&73.00±0.4&33.01±0.4&80.16±0.5    \\
\multicolumn{1}{l|}{{PLIP~\cite{huang2023visual}}}    & ResNet38  & 67.02±0.5&73.24±0.5&35.85±0.4&80.25±0.4    \\
\multicolumn{1}{l|}{}                   & ResNet50  & 66.75±0.4&72.48±0.3&32.17±0.2&80.05±0.4    \\
\multicolumn{1}{l|}{}                   & TransUNet & 61.98±0.6&66.01±0.5&21.54±0.3&76.53±0.6    \\ \hline
\multicolumn{1}{l|}{}                   & SegFormer & 66.37±0.8&72.38±0.9&32.01±0.3&80.24±0.8    \\
\multicolumn{1}{l|}{{DINOv2~\cite{oquab2024dinov2}}}  & ResNet38  & 65.01±0.6&70.47±0.6&31.90±0.4&79.23±0.6    \\
\multicolumn{1}{l|}{}                   & TransUNet & 61.00±0.4&62.71±0.5&20.22±0.2&76.64±0.4    \\ \hline
\end{tabular}
}
\caption{Performance comparison of different combinations of image encoders (CLIP, MedCLIP, PLIP, DINOv2) and backbones (SegFormer, ResNet variants, TransUNet) on initial pseudo masks for BCSS-WSSS.}
\label{tab:5}
\vspace{-0.5cm}
\end{table}

We compared our model with ten advanced methods, including three supervised models and five weakly supervised models, two state-of-the-art text-prompted weakly supervised models. We used the reported results for SSPCL and TransWS due to the unavailability of publicly available code. All other models were retrained and tested with five random seeds for robustness. Proto2Seg is a model based on human feedback; in our comparison with Proto2Seg, we carefully selected patches that met the required criteria using existing patch-level and image-level annotations instead of performing manual cutting and human feedback labelling. This might lead to a performance improvement for the Proto2Seg model.

As shown in Table~\ref{tab:1}, our PBIP framework generally outperforms state-of-the-art weakly supervised methods across the four datasets. Specifically, on BCSS-WSSS, LUAD-HistoSeg and GCSS, our method outperforms Proto2Seg by 1.68\%, 1.53\% and 1.87\% in mIoU, achieving the best weakly supervised performance. All $p$-values between our model and the second best on BCSS-WSSS and LUAD-HistoSeg are below 0.05, confirming the statistical significance of our improvements.

\begin{figure}
  \centering
    \includegraphics[width=0.48\textwidth]{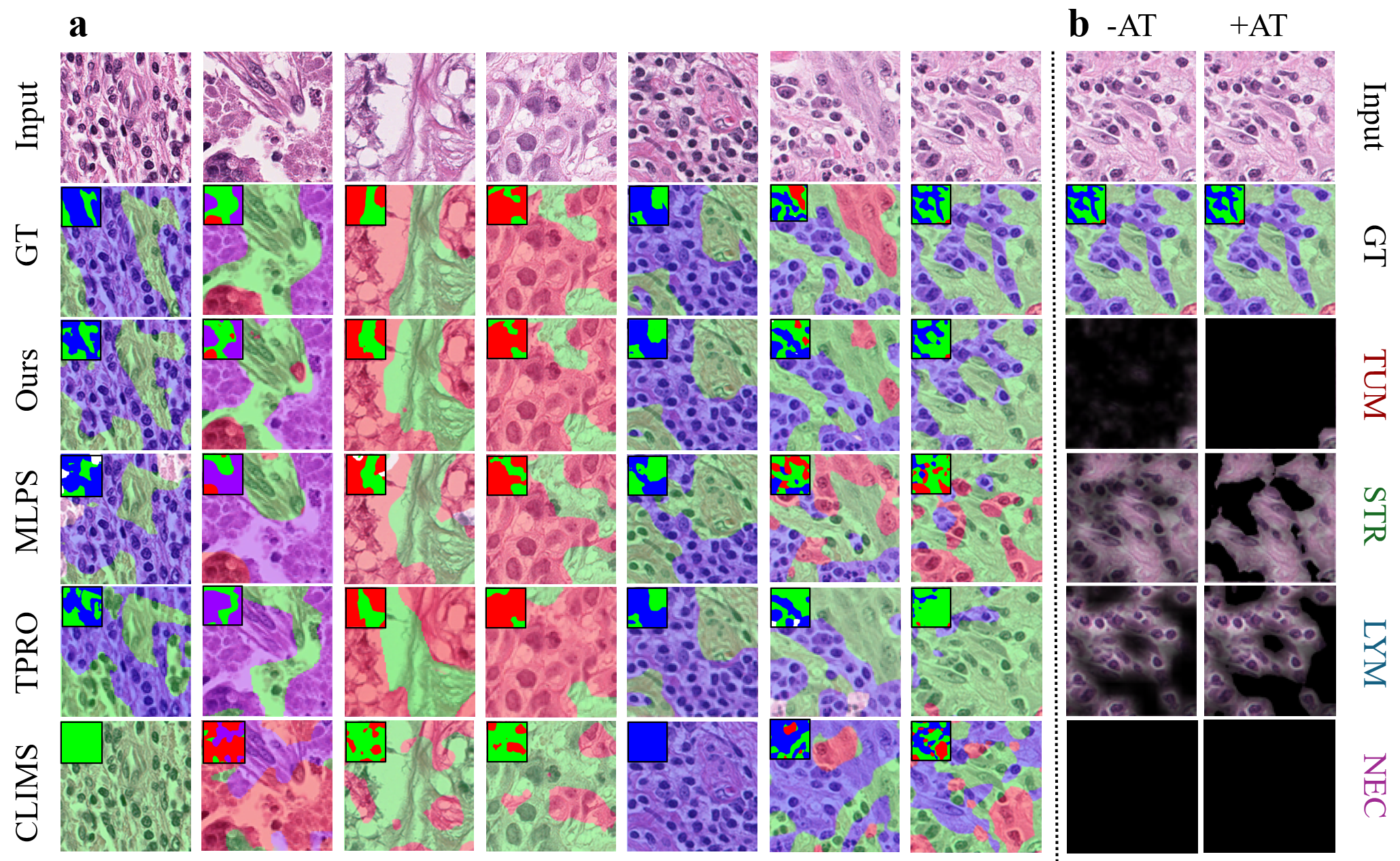}
    \caption{\textbf{a.} Visualization of the pseudo-segmentation masks generated by models in the first stage on BCSS-WSSS. The masks are overlaid on the input images, with the raw pseudo-segmentation mask shown in the top-left corner. \textbf{b.} The foreground images generated by PBIP for the four segmentation targets. More visualizations are in Supplementary Material.}
  \label{fig:visu}
  \vspace{-0.3cm}
\end{figure}

Figure~\ref{fig:visu} shows the pseudo masks generated during the first stage of WSS models. We compared PBIP with MLPS, TPRO, and CLIMS. MLPS ranks third on BCSS-WSSS, while TPRO achieves the third-best on LUAD based. CLIMS is a text-supervised segmentation model for natural images. The results indicate that PBIP activates more complete object content and fewer background regions, while TPRO offers slight improvement over MLPS and CLIMS shows many errors. Text features do not enhance histopathological segmentation, and text supervision can mislead the model due to inter-class homogeneity and intra-class heterogeneity in histopathological images.

\begin{table}[t]
\centering
\resizebox{0.48\textwidth}{!}{
\begin{tabular}{l|lllll}
\hline
K             & 1         & 2                  & 3                  & 4                  & 5         \\ \hline
Dataset       & mIoU(\%)  & mIoU(\%)           & mIoU(\%)           & mIoU(\%)           & mIoU(\%)  \\ \hline
BCSS-WSSS     & 67.03±0.4 & 67.15±0.4          & \textbf{68.01±0.6} & 67.81±0.7          & 67.53±0.7 \\
LUAD-HistoSeg & 74.91±0.5 & \textbf{75.12±0.5} & 74.87±0.6          & 73.72±1.2          & 74.07±0.8 \\
GCSS          & 51.26±0.8 & 52.07±0.8          & 53.21±1.2          & \textbf{53.91±1.0} & 53.54±0.9 \\
BCSS          & 54.01±0.6 & 53.97±0.8          & \textbf{54.05±0.8} & 53.87±1.0          & 53.50±0.9 \\ \hline
\end{tabular}
}
\caption{Ablation study on the number of clusters \( K \) in k-means across different datasets. The mIoU values on initial pseudo masks for BCSS-WSSS are reported with standard deviation. The optimal \( K \) value varies per dataset, with \( K = 2 \) to \( K = 4 \) generally achieving the highest performance.}
\label{tab:k}
\vspace{-0.4cm}
\end{table}

\begin{table}[b]
\vspace{-0.3cm}
\centering
\resizebox{0.48\textwidth}{!}{
\begin{tabular}{cc|cccc}
\hline
SIM     & AT            & mIoU(\%) & FwIoU(\%) & bIoU(\%)  & dice(\%)      \\ \hline
\checkmark    &             & 63.25±0.5& 66.80±0.4 & 25.89±0.3 & 77.98±0.4           \\
              & \checkmark  & 67.09±0.6& 73.01±0.6 & 27.48±0.5 & 79.92±0.6        \\
\checkmark    & \checkmark  &\textbf{68.01±0.6}&\textbf{73.20±0.5} &\textbf{36.88±0.4} &\textbf{81.62±0.5}\\\hline
\end{tabular}
}
\caption{Ablation study on the impact of SIM and Adaptive Thresholding (AT) on initial pseudo masks for BCSS-WSSS.}
\label{tab:modules}
\end{table}

\subsection{Ablation Studies}
\begin{figure}[t]
  \centering
    \includegraphics[width=0.48\textwidth]{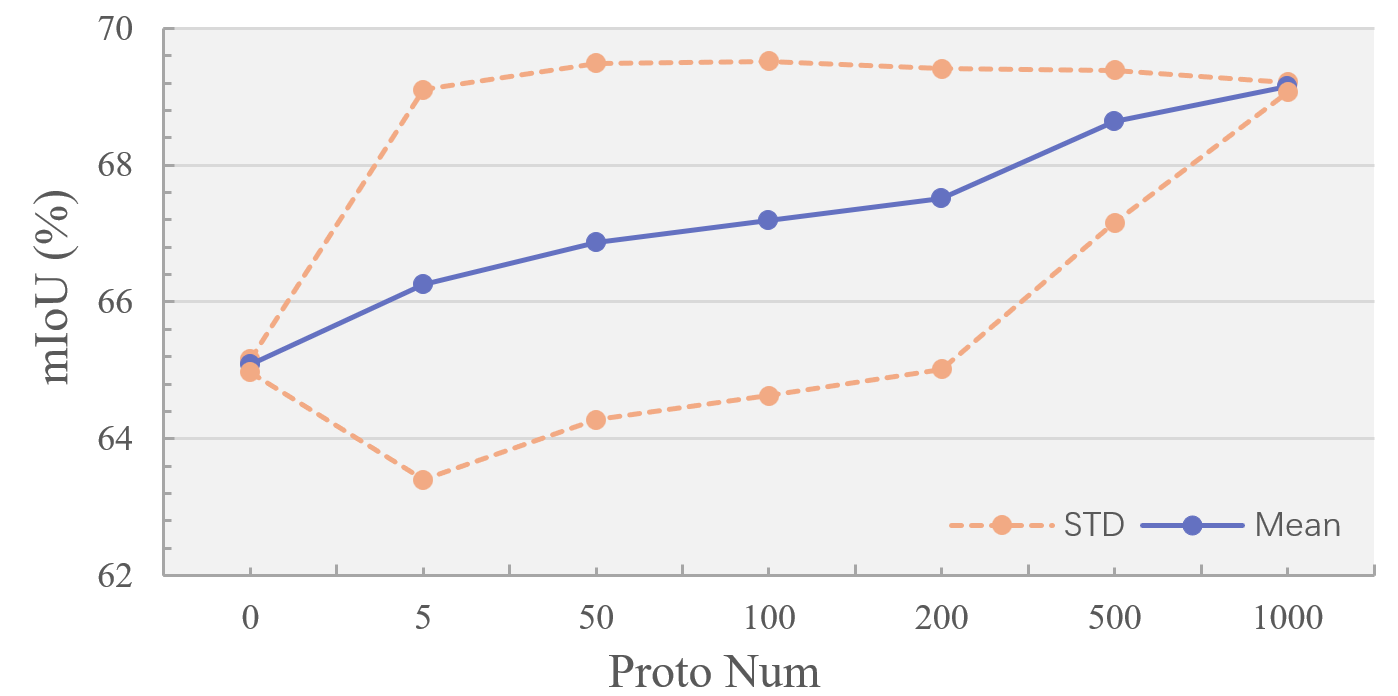}
    \caption{Ablation study of the number of prototype images. Proto Num represents the number of prototype images per sub-bank for each category. We report the mIoU values with Standard Deviation obtained over 10 runs with different random seeds. }
  \label{fig:numb}
  \vspace{-0.4cm}
\end{figure}

\textbf{Ablation study of Image Bank.} We investigated the impact of the number of prototype images by randomly selecting images without clustering, using 10 different random seeds for robustness. In Figure~\ref{fig:numb}, “Proto Num” represents the number of images per sub-bank per category. The results show that the average mIoU improves as the number of prototype images increases. Using all eligible training images yielded a mIoU of 69.15\%, suggesting that more images help reduce noise in prototype computation. However, due to inter-class homogeneity, simply increasing the number is not the most effective approach. The highest mIoU of 69.52\% was achieved with 100 images per sub-bank. When clustering was applied to select 100 images, the mIoU further increased to 69.60\%. This demonstrates that selecting representative images through clustering is more effective than merely increasing quantity, as clustering captures the most discriminative features and mitigates challenges posed by inter-class homogeneity. 

Additionally, we analyzed the impact of the number of clusters $K$ in $K$-Means. As shown in Table~\ref{tab:k}, the best results across all datasets were obtained when $K$ was set between 2 and 4, indicating that the optimal choice of $K$ may depend on the level of inter-class homogeneity and intra-class heterogeneity within the dataset. Specifically, a small \( K \) prevents the model from adequately capturing intra-class heterogeneity, and an excessively high \( K \) amplifies the influence of inter-class homogeneity.

\begin{table}[t]
\resizebox{0.48\textwidth}{!}{
\begin{tabular}{l|l|lllll}
\hline
Prompt  & Model    & TUM   & STR    & LYM    & NEC     & Mean           \\ \hline
        & CLIP~\cite{radford2021learning}     & 0.005 & 0.001  & 0.135  & 0.002   & 0.035          \\
{Text}  & PLIP~\cite{huang2023visual}     & 0.778 & 0.008  & 0.319  & 0.136   & 0.310          \\
        & MedCLIP~\cite{wang2022medclip}  & 0.423 & 0.020  & 0.165  & 0.205   & 0.203          \\ \hline
        & CLIP~\cite{radford2021learning}     & 0.716 & 0.214  & 0.529  & 0.019   & 0.370          \\
{Image} & \textbf{PLIP}~\cite{huang2023visual} & \textbf{0.924} & \textbf{0.920} & \textbf{0.816} & \textbf{0.835} & \textbf{0.874} \\
        & MedCLIP~\cite{wang2022medclip}  & 0.881 & 0.895  & 0.790  & 0.671   & 0.809          \\ \hline
\end{tabular}
}
\caption{Zero-shot classification F1 scores (\%) on the BCSS-WSSS dataset using CLIP, PLIP, and MedCLIP with text and image prompts.}
\label{tab:7}
\end{table}

\textbf{Ablation study of the Loss Function.} We analyzed the impact of different loss function configurations. Specifically, we varied the weighting ratios $\beta/\alpha$ (for $\mathcal{L}_{\text{SIM}}$ and $\mathcal{L}_{\text{CLS}}$) and $\theta_2/\theta_1$ (for $\mathcal{L}_{\text{BGS}}$ and $\mathcal{L}_{\text{FGS}}$), as illustrated in Figure~\ref{fig:hp}. The results indicate that extreme values of these ratios lead to a decline in performance. Furthermore, excluding either $\mathcal{L}_{\text{FGS}}$ or $\mathcal{L}_{\text{BGS}}$ causes the mIoU to drop significantly from 67.54\% to 50.23\% and 55.01\%, respectively. This substantial decrease suggests that both components are essential; relying solely on one causes the model to focus excessively on either the foreground or background, leading to incomplete or erroneous activations.

\textbf{Ablation study of the Modules.} We analyzed the effectiveness of different modules within  PBLP. As shown in Table~\ref{tab:modules}, both the SIM and the Adaptive Thresholding Module significantly enhance model performance. Specifically, when the SIM module is removed, we substitute it with a simple $1 \times 1$ convolutional layer to generate the pseudo-segmentation masks. The results demonstrate a noticeable decline in performance without the SIM and AT module. The SIM module captures intra-class heterogeneity and inter-class homogeneity better than the \( 1 \times 1 \) convolutional layer. Furthermore, Figure~\ref{fig:overview} illustrates the foreground images generated by the model with and without the AT module. The results demonstrate that AT effectively reduces noise in background regions.

Additionally, we evaluated the performance of our model using different image encoders and backbones, as summarized in Table~\ref{tab:5}. The combination of MedCLIP~\cite{wang2022medclip} and SegFormer achieved the best results in terms of mIoU metric. We observed that the PLIP~\cite{huang2023visual} and SegFormer pairing performed best on the bIoU, FWIoU, and Dice coefficient. The TransUNet model~\cite{chen2021transunet}, with its end-to-end architecture that does not utilize the SIM module for pseudo-mask computation, exhibited inferior performance. These results suggest that the absence of the SIM module may contribute to the lower effectiveness. Moreover, the relatively poor performance of CLIP~\cite{radford2021learning} and DINOv2~\cite{oquab2024dinov2} further highlights the importance of external knowledge from pretraining in histopathology, as these more generic encoders appear less effective for the specialized task of tissue segmentation.

\textbf{Text prompt and Image prompt.} To explore the challenges of image-text matching in histopathology, we conducted zero-shot classification experiments on the BCSS-WSSS dataset using CLIP, PLIP, and MedCLIP with both text and image prompts (see Table~\ref{tab:7}). For text prompts, we used the common format: \emph{“a photo of \{class name\}”}, while for image prompts, we utilized our prototype image features. The results show that PLIP and MedCLIP significantly outperform CLIP when using image prompts, but all models perform poorly with text prompts. This highlights the substantial challenges of text prompting in histopathology, where textual descriptions struggle to capture complex visual patterns and inter-class homogeneity. In contrast, image prompts effectively guide the models by aligning with the intricate characteristics of histopathological images through prototype features.

\section{Conclusion}
We proposed a novel Prototype-Based Image Prompting (PBIP) framework to address the challenges of inter-class homogeneity and intra-class heterogeneity in weakly supervised histopathological image segmentation. PBIP leverages image-based labels and corresponding images in histopathology datasets by extracting class prototype features through clustering methods. By incorporating class prototypes, our approach effectively mitigates inter-class homogeneity, while multiple sub-prototypes for each class address intra-class heterogeneity. Validation on four datasets demonstrated that PBIP achieves state-of-the-art performance and robustness in weakly supervised histopathological segmentation.

{
    \small
    \bibliographystyle{ieeenat_fullname}
    \bibliography{main}
}

\end{document}